\documentclass[conference,a4paper]{IEEEtran}
\IEEEoverridecommandlockouts

\usepackage[hidelinks]{hyperref}
\usepackage[cmex10]{amsmath}
\usepackage{amssymb,amsfonts}
\usepackage{dblfloatfix}

\usepackage[ruled,vlined]{algorithm2e}
\usepackage{multirow}
\usepackage{graphicx}
\usepackage[font=footnotesize,labelfont=bf]{caption}

\usepackage{subcaption}
\usepackage{booktabs}
\usepackage[export]{adjustbox}
\usepackage{float}
\graphicspath{{Figures/PDF/}{Figures/PNG/}}

\usepackage{booktabs}
\usepackage{siunitx}
\usepackage[numbers,compress]{natbib}
\usepackage{texnames}
\usepackage{bm,bbm}
\usepackage{orcidlink}
\usepackage{xcolor}

\usepackage{booktabs}
\usepackage{geometry}
\usepackage{array}
\usepackage{graphicx}
\begin{document}

\title{SDHSI-Net: Learning Better Representations for Hyperspectral Images via Self-Distillation}

\author{
Prachet Dev Singh\textsuperscript{1}\orcidlink{0009-0001-0983-2378}, 
Shyamsundar Paramasivam\textsuperscript{1}\orcidlink{0009-0003-9003-8263}, 
Sneha Barman\textsuperscript{2}\orcidlink{0009-0001-4711-5695},\\
Mainak Singha\textsuperscript{3}\orcidlink{0000-0002-7615-2575}, 
Ankit Jha\textsuperscript{1}\orcidlink{0000-0002-1063-8978}, 
Girish Mishra\textsuperscript{4}\orcidlink{0000-0002-7489-2721}, 
Biplab Banerjee\textsuperscript{3}\orcidlink{0000-0001-8371-8138} \\
\textsuperscript{1}\textit{Dept. of CSE, The LNMIIT Jaipur} \quad
\textsuperscript{2}\textit{DAGP, IIT Dhanbad} \quad
\textsuperscript{3}\textit{CSRE, IIT Bombay} \quad
\textsuperscript{4}\textit{SAG, DRDO, Delhi}
}

\maketitle
\begin{abstract}
Hyperspectral image (HSI) classification presents unique challenges due to its high spectral dimensionality and limited labeled data. Traditional deep learning models often suffer from overfitting and high computational costs. Self-distillation (SD), a variant of knowledge distillation where a network learns from its own predictions, has recently emerged as a promising strategy to enhance model performance without requiring external teacher networks. In this work, we explore the application of SD to HSI by treating earlier outputs as soft targets, thereby enforcing consistency between intermediate and final predictions. This process improves intra-class compactness and inter-class separability in the learned feature space. Our approach is validated on two benchmark HSI datasets and demonstrates significant improvements in classification accuracy and robustness, highlighting the effectiveness of SD for spectral-spatial learning. Codes are available at \url{https://github.com/Prachet-Dev-Singh/SDHSI}.
\end{abstract}


\section{Introduction}

Hyperspectral imaging (HSI) is an advanced remote sensing technology that collects detailed spectral information across hundreds of narrow and contiguous wavelength bands. Unlike traditional RGB or multispectral imaging systems that capture only a few broad bands, HSI provides a complete spectral signature for each pixel in a scene. This allows for fine-grained material identification, making it especially useful in domains where materials may appear visually similar but exhibit distinct spectral properties. Applications of HSI span across precision agriculture (e.g., crop disease detection and nutrient monitoring), mineral and geological exploration, urban land-cover mapping, and environmental monitoring, including forest health assessment and pollution detection \cite{Grewal2023MachineLA, Haidar2024EnhancingHI}. The ability to capture such detailed spectral profiles makes HSI particularly suitable for tasks requiring subtle differentiation between surface materials. However, this wealth of information also introduces significant computational and modeling challenges due to the high dimensionality, spectral redundancy, and often limited availability of labeled training samples.

Earlier methods for hyperspectral image classification primarily utilized classical machine learning algorithms such as Support Vector Machines (SVMs) and Random Forests \cite{9206124}. These techniques depended heavily on handcrafted features that attempted to summarize the spectral and spatial characteristics of each pixel. While effective in controlled settings, these features often lacked generalizability and failed to scale across datasets with varying sensor characteristics or environmental conditions. With the emergence of deep learning, especially convolutional neural networks (CNNs), feature extraction became data-driven and more robust. CNN-based models, including 3D-CNNs \cite{article, mdfs-net} and spectral-spatial attention networks, demonstrated improved classification performance by learning joint representations of spectral and spatial dimensions \cite{Lin2013SpectralspatialCO}. Despite these advances, most deep learning models require large computational resources and substantial labeled data, making their deployment on embedded systems such as UAVs or on-board satellite processors less practical \cite{Sigger2023DiffSpectralNetU}.

To bridge this gap, researchers have turned to model compression techniques like pruning, quantization, and knowledge distillation (KD) to reduce model size and computational load. Among these, KD has shown considerable success by transferring the learned knowledge from a large teacher network to a smaller student network \cite{hinton2015distillation}. This student network then mimics the outputs of the teacher during training, often achieving similar performance with reduced complexity. However, KD assumes access to a fully trained, high-capacity teacher model, which is often infeasible in hyperspectral settings where labeled data is scarce and training deep models is computationally expensive. Furthermore, many conventional KD methods overlook the distinct structure and redundancy patterns of hyperspectral data, limiting their transferability and impact.

In this paper, we present Self-distillation for Hyperspectral Image classification (SDHSI-Net) which is a lightweight deep learning network inspired from \cite{szegedy2014goingdeeperconvolutions}. Also in order to improve feature learning, employ a self-distillation (SD) strategy \cite{zhang2019teacherimproveperformanceconvolutional, DBLP:conf/bmvc/JhaKBN20} designed specifically for HSI. Instead of relying on a separate teacher, our model transfers knowledge internally between intermediate and deeper layers\cite{zhang2019teacherimproveperformanceconvolutional}, enabling efficient representation learning without external supervision\cite{Wu2025OvercomingGM}.
Our main contribution lies in combining multi-level SD with a triplet loss objective [10], creating a unified strategy that enhances class separability and improves generalization under limited data. This integration allows the network to refine its spectral–spatial features and learn more discriminative embeddings while maintaining low computational cost.
We summarize our key contributions as:\\
\noindent -We propose a hierarchical self-distillation framework named SDHSI-Net for hyper-spectral image classification, which refines the internal features using distillation loss bewteen the teacher and student modules.\\
\noindent -We also introduce triplet loss during training to enhance class separability by structuring the feature space in a way that better distinguishes between spectrally similar classes.\\
\noindent -We compare the performance of our proposed SDHSI-Net against state-of-the-art (SOTA) HSI classification methods on the two benchmark datasets.

\section{Methodology and Proposed Model}
\label{sec:method_proposed_model}

\subsection{Preliminaries}

We aim to learn a parametric function that maps a hyperspectral image to its corresponding semantic label space by leveraging both spectral and spatial information. Let the hyperspectral dataset be denoted as \( D = \{(\mathbf{I}, Y)\} \), where \( \mathbf{I} \in \mathbb{R}^{H \times W \times C} \) is the input image, and \( Y \in \mathbb{R}^{H \times W} \) represents the ground-truth label map. Here, \( H \) and \( W \) denote the spatial dimensions, and \( C \) is the number of spectral bands. To address the high spectral dimensionality and reduce computational overhead, we apply Principal Component Analysis (PCA) along the spectral dimension. This projects the input data onto a lower-dimensional subspace, yielding a transformed image \( \mathbf{I}_{\text{PCA}} \in \mathbb{R}^{H \times W \times B} \), where \( B \ll C \) is the number of retained principal components. From \( \mathbf{I}_{\text{PCA}} \), we extract overlapping spatial-spectral patches of size \( S \times S \times B \), centered at each labeled pixel. These 3D patches are used as inputs to our model, allowing it to learn localized spectral-spatial representations that are crucial for accurate classification.

\subsection{Our proposed SDHSI-Net}
Self-Distilled Hyperspectral Image Network (SDHSI-Net) is a lightweight hybrid architecture designed for hyperspectral image classification. It consists of a shared 3D-2D convolutional backbone followed by fully connected layers. Within SDHSI-Net, we define the deeper branch of the network as the \textit{Teacher module}, and introduce two auxiliary, shallower branches, referred to as \textit{Student modules}, which are attached at intermediate layers of the network. These Student modules enable multi-level self-distillation during training through distillation-based loss functions. In addition, the model incorporates a metric learning objective using triplet loss to enhance the discriminative quality of the learned feature embeddings. In the following subsections, we first describe the architecture of the Teacher and Student modules, and then detail the network optimization process.

\noindent\textbf{Teacher module:}  
The Teacher module serves as the primary classification backbone of SDHSI-Net and is responsible for learning deep spectral-spatial representations from hyperspectral data. It takes as input a spatial-spectral patch extracted from the PCA-compressed hyperspectral image. To effectively capture both spectral and spatial dependencies, the module begins with a series of 3D convolutional layers that jointly process information across spatial dimensions and spectral bands. Each convolutional layer is followed by normalization and a non-linear activation function to ensure stable training and enhance representational capacity. The output of the 3D block is then reshaped to merge the spectral and channel dimensions, enabling further refinement through a sequence of 2D convolutional layers focused on spatial learning. To promote generalization and robustness, regularization techniques such as DropBlock are employed. Finally, the refined features are passed through a fully connected classification head with dropout, which produces the final class probabilities.

\begin{figure}
\centering
\includegraphics[width=\linewidth]{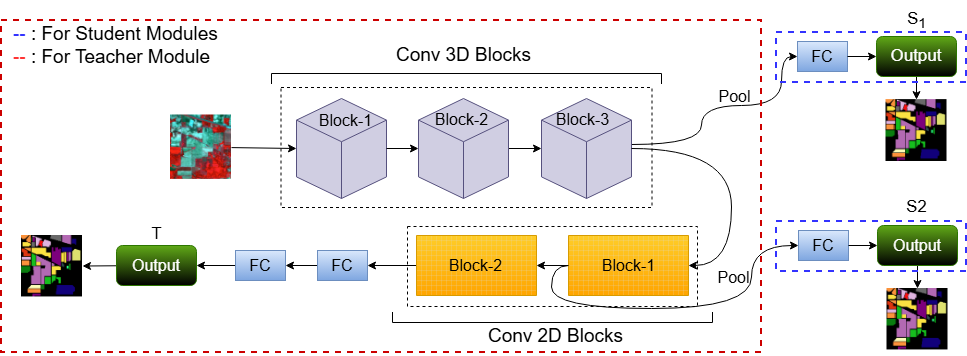}
\caption{The architecture diagram of our SDHSI-Net, where $T$, $S_1$, and $S_2$ are the outputs from the Teacher, Student 1 and Student 2 modules, respectively. We denote the fully connected layer with FC. The distillation takes place between the Teacher $T$ and Students ($S_1$ and $S_2$) using the logit distillation and hint distillation losses, described in Section \ref{sec:method_proposed_model}. Also, we incorporated the triplet loss for better separability of classes.}
\label{fig:model_architecture}
\end{figure}
\noindent\textbf{Student module:} To enable multi-level supervision while keeping inference efficient, SDHSI-Net integrates two lightweight Student modules that assist the Teacher module during training. These Student modules are designed to mimic the Teacher’s outputs at different intermediate layers, promoting hierarchical feature refinement through self-distillation. Empirically, we observed that integrating two Student heads provided improved performance over other setups. Specifically, the first Student module branches off after the final 3D convolutional layer, before reshaping, and learns from the intermediate spatio-spectral representations. The second Student module is attached to the output of the last 2D convolutional layer, just before the classifier, and focuses on refined high-level spatial features. Each Student module consists of a compact structure comprising a pooling layer followed by a lightweight multi-layer perceptron (MLP) classifier. Importantly, these modules are only active during training, ensuring that the additional computations introduced for distillation do not affect inference-time performance. This design aligns with the core motivation of SDHSI-Net: to achieve classification performance comparable to larger models while significantly reducing computational cost and inference latency. An overview of the complete architecture, including the Teacher and Student modules, is illustrated in Figure~\ref{fig:model_architecture}.

\noindent\textbf{Self-distillation and triplet learning:} SDHSI-Net takes input of PCA-compressed spatial-spectral patches using a hybrid 3D-2D backbone to learn joint representations. To improve feature learning without increasing inference cost, it incorporates a multi-level SD framework with two lightweight Student modules that mimic intermediate representations of the Teacher network. This hierarchical supervision stabilizes early layers and enhances generalization. Training supervision combines cross-entropy loss with logit-level and feature-level distillation, all performed online without requiring a pre-trained teacher. Additionally, a triplet loss over feature embeddings promotes intra-class compactness and inter-class separability, enabling the network to learn robust and discriminative features even under limited data conditions.

\subsection{Network optimization}

We train our proposed SDHSI-Net using composite loss function that combines supervision at the output and intermediate levels. This includes standard classification loss, SD objectives, and a metric-based triplet loss to enhance discriminative power.\\
\noindent\textbf{Cross-Entropy loss:}  
We train the classification heads used in the teacher module and student modules using the cross-entropy loss which is defined in Equation \ref{eq:teacher}, where $y_i$ is the one-hot encoded ground-truth label, $\log(\hat{y}_{T_i})$ is the teacher's output and $\log(\hat{y}_{S_i})$ is output from the respective student modules and $K$ is the total number of classes.
\begin{equation}\label{eq:teacher}
\mathcal{L}_{\text{CE}}^{T} = - \sum_{i=1}^{K} y_i \log(\hat{y}_{T_i}), \quad\mathcal{L}_{\text{CE}}^{S} = - \sum_{i=1}^{K} y_i \log(\hat{y}_{S_i})
\end{equation}
\noindent\textbf{Distillation losses}  
To align intermediate and output-level predictions of the Student modules with the Teacher, we introduce two self-distillation objectives:
\begin{itemize}
\item \textbf{Logit distillation loss ($\mathcal{L}_{\text{Logit}}$)}:
The motivation of this loss is to match the output of the Student's soft predictions to match those of the Teacher module, which consists of  the mean squared error (MSE) between the logits of corresponding modules.
\begin{equation}
\mathcal{L}_{\text{Logit}} = \frac{1}{N} \sum_{i=1}^{N} \left\| z_i^{\text{T}} - z_i^{\text{S}} \right\|_2^2
\end{equation}
where $z_i^{\text{T}}$ and $z_i^{\text{S}}$ are the logits from the Teacher and Student, respectively.

\item \textbf{Hint loss ($\mathcal{L}_{\text{Hint}}$)}:  
To align internal feature representations, we apply L2 loss between normalized intermediate features from the Student and Teacher, which helps the shallow Student modules approximate high-level semantic features from the Teacher.
\begin{equation}
\mathcal{L}_{\text{Hint}} = \frac{1}{N} \sum_{i=1}^{N} \left\| \frac{f_i^{\text{T}}}{\|f_i^{\text{T}}\|} - \frac{f_i^{\text{S}}}{\|f_i^{\text{S}}\|} \right\|_2^2
\end{equation}
\end{itemize}

\noindent\textbf{Triplet loss:}  
To promote better separation between classes in the learned feature space, we incorporate a triplet loss over sample-level feature embeddings. For each labeled training sample, a triplet is formed consisting of an anchor, a positive, and a negative. The anchor is the feature embedding of a given HSI input sample, the positive is the embedding of another sample from the same class, and the negative is the embedding of a sample from a different class, selected via hard mining within the mini-batch. These embeddings are extracted from the Teacher network. The triplet loss encourages the anchor to be closer to the positive than to the negative by at least a margin \( m \), thereby enforcing intra-class compactness and inter-class separability in the embedding space. It is defined as:
\begin{equation}
\mathcal{L}_{\text{Trip}} = \frac{1}{N} \sum_{k=1}^{N} \max \left( 0, \|A_k - P_k\|_2^2 - \|A_k - N_k\|_2^2 + m \right)
\end{equation}
\noindent where \( A_k \), \( P_k \), and \( N_k \) denote the anchor, positive, and negative embeddings for the \( k \)-th triplet, respectively. This objective leads to more discriminative and generalizable feature representations, especially beneficial in hyperspectral classification tasks where inter-class spectral similarities are common.\\
\noindent\textbf{Total loss:}  
We train SDHSI-Net using a composite loss function that integrates multiple learning objectives. The overall loss is defined as:
\begin{equation}
\mathcal{L}_{\text{total}} = \mathcal{L}_{\text{CE}}^{\text{T}} + \mathcal{L}_{\text{CE}}^{\text{S}} + \lambda_{\text{L}} \mathcal{L}_{\text{Logit}} + \lambda_{\text{H}} \mathcal{L}_{\text{Hint}} + \lambda_{\text{trip}} \mathcal{L}_{\text{Trip}}
\end{equation}

\noindent
where \( \lambda_{\text{L}} \), \( \lambda_{\text{H}} \), and \( \lambda_{\text{triplet}} \) denote the weighting coefficients for the logit distillation loss, hint loss, and triplet loss, respectively. The proposed SDHSI-Net not only improves prediction accuracy but also facilitates learning of more stable, discriminative, and transferable internal representations. This joint optimization strategy enhances the model’s robustness in low-data hyperspectral scenarios.
 

\section{Experimental Evaluation}
\label{sec:exp}
\noindent\textbf{Datasets description:} We evaluate our model on two standard HSI datasets. a) \textbf{Indian Pines (IP)} covers a $145 \times 145$ region in Indiana with 220 spectral bands, of which 200 are retained after removing noisy bands and it includes 16 land cover classes. b) \textbf{Salinas (SA)} spans $512 \times 217$ pixels over agricultural fields in California containing processed 204 spectral channels with 16 classes. We applied PCA to retain the top $30$ components for both datasets. Also, we extracted 3D patches of size $17 \times 17 \times B$, where $B=30$.  \\
\noindent\textbf{Architecture details and training protocols:} Our model follows the architecture detailed in Section~\ref{sec:method_proposed_model}, the backbone comprises of four 3D convolutional layers and two 2D convolutional layers, followed by two fully connected layers. To enable self-distillation, we integrate two auxiliary student heads - each a 2-layer MLP with 128 hidden units - after the final 3D and initial 2D CNN stages. We also incorporate the self-attention mechanism in the student modules (shallower heads) and the final output head acts as the teacher module.
\begin{table}[t]
\centering
\caption{Comparison of SDHSI-Net against SOTA methods. Best and second best results depicted in \textbf{bold} and \textcolor{blue}{blue}, respectively.}
\label{tab:results_indian_salinas}
\scalebox{0.94}{
\begin{tabular}{lcccccc}
\hline
\textbf{Methods}               & \multicolumn{3}{c}{\textbf{Indian Pines Dataset}} & \multicolumn{3}{c}{\textbf{Salinas Scene Dataset}} \\
\cline{2-4} \cline{5-7}
                               & \textbf{OA}        & \textbf{\(\kappa\)}& \textbf{AA}        & \textbf{OA}        & \textbf{\(\kappa\)}     & \textbf{AA}        \\
\hline

2D-CNN \cite{article-1}        & 89.48              & 87.96              & 86.14              & 97.38              & 97.08              & 98.84              \\
3D-CNN \cite{article}          & 91.10              & 89.98              & 91.58              & 93.96              & 93.32              & 97.01              \\
M3D-CNN \cite{inproceedings-1} & 95.32              & 94.70              & 96.41              & 94.79              & 94.20              & 96.25              \\
SSRN \cite{article-2}          & \textbf{99.19}     & \textbf{99.07}     & \textbf{98.93}     & \textbf{99.98}     & \textcolor{blue}{99.97}     & \textbf{99.97}     \\
HybridSN\cite{Roy_2020}        & 98.54              & 98.35              & 97.62              & 99.95              & 99.95              & \textbf{99.97}     \\
GSC-VIT\cite{10472541} & 97.12 & 96.67 & 94.34 & 97.15 & 96.88 & 97.99 \\

IGSSMamba\cite{he2024igroupss} & 98.71 & 98.53 & 98.33 & - & - & - \\
\hline
{{SDHSI-Net ($T$)}}            &\textcolor{blue}{99.00} & \textcolor{blue}{98.86} & \textcolor{blue}{98.71} & {\textbf{{99.98}}} & {\textbf{{99.98}}} & {\textbf{99.97}} \\
{{SDHSI-Net ($S_1$)}}          & {{93.64}} & {{92.75}} & {{74.87}} & {{99.79}} & {{99.77}} & \textcolor{blue}{99.84} \\
{SDHSI-Net ($S_2$)}            & {{98.65}} & {{98.47}} & {{94.95}} & {\textcolor{blue}{{99.96}}} & {{99.96}} & {\textbf{99.97}} \\
\hline
\end{tabular}
}
\end{table}

\textbf{Training and evaluation protocol:} We have trained SDHSI-Net using the AdamW optimizer with an initial learning rate of $5 \times 10^{-4}$, decayed to $1 \times 10^{-6}$ via cosine annealing. A weight decay of $1 \times 10^{-5}$ was applied. All models have been trained for 100 epochs with a batch size of 64, without early stopping. The loss weight coefficients are empirically selected as $\lambda_{\text{CE}} = 1.0$, $\lambda_{\text{L}} = 1 \times 10^{-5}$, $\lambda_{\text{H}} = 0.001$, and $\lambda_{\text{trip}} = 0.001$, with the triplet margin set to $m = 0.2$. We evaluate the model performance using Overall Accuracy (OA), Average Accuracy (AA), and Cohen’s Kappa Coefficient ($\kappa$).

\subsection{Comparison to the literature}

We evaluate our SDHSI-Net model against the SOTA methods on the Indian Pines and Salinas datasets. While comparing with SVM, the Teacher, Student heads $S_1$ and $S_2$ outperform OA by 14.7\%, 8.34\% and 13.35\%, respectively on Indian Pines. Furthermore, we compare SDHSI-Net with 2D-CNN \cite{article-1}, 3D-CNN \cite{article}, M3D-CNN \cite{inproceedings-1}, SSRN \cite{article-2} and HybridSN \cite{Roy_2020}, where we have observed that the teacher head obtained comparable results against SSRN and outperforms HybridSN averagely on both the dataset by $0.25\%$, where as the Student head $S_2$ surpasses OA metric against 2D-CNN, 3D-CNN, M3D-CNN and HybridSN by $9.17\%$, $7.55\%$, $3.33\%$ and $0.11\%$ on Indian Pines dataset. We observe the similar pattern on the Salians dataset where SDHSI-Net performs nearly similary to the best SOTA method i.e., SSRN \cite{article-2}.

\begin{table}[ht]
\centering
\caption{Performance comparison of SDHSI-Net for with (w) and without (w/o) SD, training configuration, number of training parameters and inference time on the Salinas dataset.}
\label{tab:self_distill}
\scalebox{0.85}{
\begin{tabular}{lcc|ccc|c|c}
\toprule
\multirow{2}{*}{\textbf{Head}}&\multicolumn{2}{c|}{\textbf{Distillation}}&\multicolumn{3}{c|}{\textbf{Train-val-test}}&{\textbf{\#P}}&{\textbf{Time}}\\\cmidrule(lr){2-6}

& \textbf{w/o SD}& \textbf{w SD}&\textbf{C1}&\textbf{C2}&\textbf{C3}& (M)&($\mu$s)\\
\midrule
$S_1$        & 3.33& 99.79&99.79&98.87&99.39&0.11&0.89\\
$S_2$        & 2.28& 99.96&99.96&99.84&99.85&1.30&1.43\\
Teacher  & 99.91& \textbf{99.98}&\textbf{99.98}&\textbf{99.93}&\textbf{99.90}&1.70&1.64\\
\bottomrule
\end{tabular}}
\end{table}

\subsection{Ablation studies}
\noindent{\textbf{Comparing model with and without SD:}}
To evaluate the impact of hierarchical SD, we compare model performance with and without it. As shown in Table~\ref{tab:self_distill}, SD boosts intermediate student heads achieves comparable performance against teacher head, bringing them close to teacher-level performance. Despite using fewer parameters and requiring less inference time, the student heads still retain strong discriminative power, showing the effectiveness of SD.\\
\noindent{\textbf{Effect of varying training split:}} We ablate the SDHSI-Net with different training-validation-testing splits across three configurations; C1 = 30-10-60, C2 = 20-10-70, and C3 = 10-10-80. The performance of each configuration was measured across two student heads ($S_1$ and $S_2$) and one final teacher head and results are shown in Table~\ref{tab:self_distill}. As observed, all components show a consistent trend of performance degradation as the training data is reduced. While Student $S_2$ and the Teacher Final Head maintain relatively high accuracies across all splits, Student Head 1 is more sensitive to the decrease in training data, with a noticeable drop of 2.3\%  as the training split changes from 30\% to 10\%.  The results indicate that the model remains robust even with lower training data, especially for Student $S_2$ and the Teacher model. However, maintaining a higher proportion of training data yields the most optimal and consistent performance across all heads.\\
\noindent{\textbf{Number of training parameters:}} With self-distillation, student heads achieve near-teacher performance while using far fewer parameters. As shown in Table \ref{tab:self_distill}, $S_1$ uses 10\% of the teacher’s parameters and still exceeds 99\% precision. $S_2$, with about 25\% fewer parameters, slightly outperforms the teacher. This shows that our self-distilled design enables efficient learning with compact models across all train-test splits.\\
\noindent{\textbf{Impact of $\mathcal{L}_{trip}$}:}
To assess the contribution of triplet loss, we have evaluated the model on the Indian Pines dataset. As shown in \ref{tab:triplet_loss_comparison}, the incorporation of triplet loss consistently improves performance across all metrics. The improvement being: incorporating triplet loss improved the final head performance, increasing overall accuracy by 0.2\% and average accuracy by 0.97\%.\\

\begin{table}[ht]
\centering
\caption{Comparison of final head metrics with and without triplet loss on the Indian Pines dataset.}
\label{tab:triplet_loss_comparison}
\scalebox{0.9}{%
\begin{tabular}{lcc}
\toprule
\textbf{Metric} & \textbf{With Triplet Loss} & \textbf{Without Triplet Loss} \\
\midrule
OA & 98.93\% & 98.73\% \\
AA & 98.23\% & 97.26\% \\
$\kappa$ & 0.9878 & 0.9855 \\
\bottomrule
\end{tabular}
}
\end{table}

\noindent{\textbf{Effect on varying patch sizes:}} In \ref{tab:patch_head_perf} we quantitatively observe that increasing the patch size leads to improvement in model performance. We observed that with the increasing the patch size from 11×11 to 17×17 improved performance across all heads, with the Teacher module ($T$) achieving the highest gains: 26.33\% in average accuracy and 3.72\% in overall accuracy. So it can be concluded that 17x17 is the optimal patch for classification.\\

\begin{table}[ht]
\centering
\caption{Comparison of classification performance across patch sizes on the Indian Pines dataset.}
\label{tab:patch_head_perf}
\scalebox{0.85}{%
\begin{tabular}{clccc}
\toprule
\textbf{Patch Size} & \textbf{Metric} & \textbf{$S_1$} & \textbf{$S_2$} & \textbf{Teacher} \\
\midrule
\multirow{3}{*}{11$\times$11}
& OA        & 79.36  & 96.53  & 95.28 \\
& AA        & 50.94  & 82.34  & 72.38 \\
& $\kappa$  & 0.7617 & 0.9605 & 0.9462 \\
\midrule
\multirow{3}{*}{15$\times$15}
& OA        & 93.69  & 98.64  & 98.86 \\
& AA        & 74.96  & 92.72  & 97.27 \\
& $\kappa$  & 0.9280 & 0.9846 & 0.9871 \\
\midrule
\multirow{3}{*}{17$\times$17}
& OA        & 93.64  & 98.65  & 99.00 \\
& AA        & 74.87  & 94.95  & 98.71 \\
& $\kappa$  & 0.9275 & 0.9847 & 0.9886 \\
\bottomrule
\end{tabular}
}
\end{table}
\vspace{-0.4em} 

\begin{figure}[ht!]
\begin{center}
\includegraphics[width=\columnwidth]{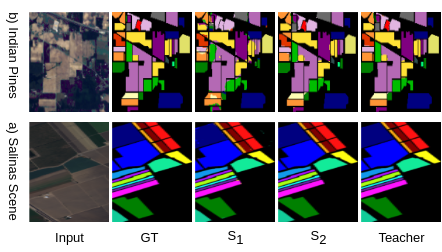}
\caption{The classification maps obtained from our proposed model SDHSI-Net on a) Indian Pines and b) Salinas datasets. Here, GT, $S_1$, $S_2$ and $T$ denote the ground truth, Student 1, Student 2 and Teacher, respectively.}
\label{fig:salinas_indian_comparison}

\end{center}
\vspace{-0.5cm}
\end{figure}
\noindent\textbf{Model output visualization:}
Figure~\ref{fig:salinas_indian_comparison} shows segmentation results on the Indian Pines and Salinas datasets. Each row includes the input image, ground truth (GT), and predictions from Student Heads ($S_1$, $S_2$) and the Teacher (T). All predictions are visualized as semantic segmentation maps, with each color representing a land cover class. Both student heads closely match the GT, with $S_2$ showing finer detail due to its deeper position. The teacher output is slightly more refined, confirming that self-distillation enables strong intermediate supervision.

\section{Conclusion and Future Work}
In this paper, we have proposed SDHSI-Net, a lightweight and effective model for HSI classification that remains robust even with limited training data and high spectral dimensionality. By integrating a multi-level SD mechanism, SDHSI-Net enhances the learning capacity of both intermediate and final layers, leading to improved classification performance. This design also enables the optional use of earlier student heads during inference, offering a trade-off between accuracy and efficiency, making it suitable for resource-constrained applications. We compared SDHSI-Net with several SOTA methods and conducted comprehensive ablation studies to demonstrate the effectiveness of our self-distillation framework. Notably, the student modules alone achieve competitive performance while significantly reducing inference time and computational cost. For future work, we plan to extend SDHSI-Net to more complex hyperspectral tasks such as change detection and semantic segmentation, where efficient and discriminative feature learning remains a key challenge.

\small
\bibliographystyle{IEEEtranN}
\bibliography{references}

\end{document}